\documentclass{article}
\usepackage[preprint]{neurips_2020}

\usepackage[small]{caption}
\usepackage[utf8]{inputenc} % allow utf-8 input
\usepackage[T1]{fontenc}    % use 8-bit T1 fonts
\usepackage{hyperref}       % hyperlinks
\usepackage{url}            % simple URL typesetting
\usepackage{booktabs}       % professional-quality tables
\usepackage{amsfonts}       % blackboard math symbols
\usepackage{nicefrac}       % compact symbols for 1/2, etc.
\usepackage{microtype}      % microtypography
\usepackage[dvipsnames]{xcolor}
\usepackage{amsmath}
\usepackage{upgreek}
\usepackage{graphicx}
\usepackage{algpseudocode,algorithm}
\usepackage{algorithmicx}
\usepackage{multirow}
\usepackage{caption}
\usepackage{subcaption}
\usepackage{stfloats}
\usepackage{comment}
\usepackage{tabularx}

\setcitestyle{numbers,open={[},close={]},citesep={,}}

\DeclareMathOperator*{\argmax}{argmax}
\newcommand{\tabitem}{~~\llap{\textbullet}~~}

\title{Reinforced Data Sampling for \\ Model Diversification}

\author{%
  Hoang D. Nguyen \\
  School of Computing Science \\
  University of Glasgow, Singapore \\
  \texttt{Harry.Nguyen@glasgow.ac.uk} \\
  \And
  Xuan-Son Vu \\
  Department of Computing Science \\
  Ume\r{a} University, Sweden \\
  \texttt{sonvx@cs.umu.se} \\
  \AND
  Quoc-Tuan Truong \\
  School of Information Systems \\
  Singapore Management University, Singapore \\
  \texttt{qttruong.2017@smu.edu.sg} \\
  \And
  Duc-Trong Le \\
  University of Engineering and Technology \\
  Vietnam National University, Vietnam \\
  \texttt{trongld@vnu.edu.vn} \\
}

\begin{document}

\maketitle

\begin{abstract}
With the rising number of machine learning competitions, the world has witnessed an exciting race for the best algorithms. However, the involved data selection process may fundamentally suffer from evidence ambiguity and concept drift issues, thereby possibly leading to deleterious effects on the performance of various models. This paper proposes a new Reinforced Data Sampling (RDS) method to learn how to sample data adequately on the search for useful models and insights. We formulate the optimisation problem of model diversification $\delta{-div}$ in data sampling to maximise learning potentials and optimum allocation by injecting model diversity. This work advocates the employment of diverse base learners as value functions such as neural networks, decision trees, or logistic regressions to reinforce the selection process of data subsets with multi-modal belief. We introduce different ensemble reward mechanisms, including soft voting and stochastic choice to approximate optimal sampling policy. The evaluation conducted on four datasets evidently highlights the benefits of using RDS method over traditional sampling approaches. Our experimental results suggest that the trainable sampling for model diversification is useful for competition organisers, researchers, or even starters to pursue full potentials of various machine learning tasks such as classification and regression. The source code is available at \url{https://github.com/probeu/RDS}.
\end{abstract}

\section{Introduction}

Data sampling is the process of selecting subsets of data points for analysis and reporting in the larger dataset. In machine learning, it is a fundamental step to ensure learning methods to generalise new observations adequately. However, classical data sampling techniques like randomisation are susceptible to detrimental issues on learning performance, including concept drift and evidence ambiguity \cite{bach2010bayesian}. Consider a machine learning task such as classification or regression, in which it aims at modeling $f: x \to y$ using training data samples in light of maximising its task performance. The model concept, or mappings from data observations to outputs, may drift rendering sub-optimal fit with new data points due to hidden context changes or data-related problems. Often, there is simply insufficient or inappropriate information presented in the data samples to support adequate predictive strength, which is also known as evidence ambiguity \cite{lee2016stochastic}. Hence, the nexus between data points and model outputs plays a crucial role in data selection to improve generalisation and to mitigate such issues \cite{zhou2018diverse}. This paper describes a new sampling method, named as Reinforced Data Sampling (RDS), to learn how to sample data effectively predicated on base models in searching for useful models and insights. We focus on the ensemble use of multiple models to enhance the representational ability of the data and to select subsets of data points according to future learning potentials \cite{gong2019diversity}.

Models are different in their strengths and weaknesses, thus exploiting their disagreements is a useful mechanism for better learning performance. Sample-based model diversification has been proven as an effective ensemble strategy in machine learning \cite{carreira2016ensemble}. Given $K$ learners $\mathcal{F} = \{f_1, f_2, ..., f_K\}$, diverse ensembles $f_\epsilon$ can be derived from the collective behaviours of the members $f_k \in \mathcal{F}$ \cite{lee2016stochastic}. As typical learners are apt to mode-seeking behaviours, sample-based randomisation can be used to inject model diversity for gaining more complement information from multiple learners \cite{carreira2016ensemble}. Therefore, learning how to sample with the consideration of task performance and model diversity is intriguing to support many applications in machine learning.

In this paper, we formulate a sampling problem for model diversification $\delta{-div}$, in which a data sampler is trained to generate subsets of the larger dataset. Moreover, we advocate the use of diverse learners $f_k$ in our method to promote model diversity by design, including stable learning methods (e.g., support vector machines, regularised least square regression) and unstable learning methods (e.g., neural networks or decision trees) \cite{bach2008paired,minku2009impact}. Our approach entails reinforcement learning of observable evidence in the dataset to approximate the parameters of our sampler with ensemble value functions. We propose several novel ensemble reward mechanisms, collectively soft voting and stochastic choice. Furthermore, our method is designed to achieve proportionate allocation with regularisation of distributional property.

We evaluate the RDS approach using four datasets, including NIPS 2003 Feature Selection Challenge - Madelon, Kaggle Hackaton - Drug Reviews, MNIST, and Kalapa Credit Scoring Challenge. Our experiments cover a range of machine learning tasks such as binary classification, multi-class classification, and regression on multivariate, textual, and visual data. The results evidently highlight better performance impacts of trainable data samples over classical or prior data selection.

In data challenges, AI and data scientists have witnessed the deleterious effects of sub-optimal data preparation in large-scale competitions. It entangles the machine learning community with limitations for analysis and reporting, thereby restricting innovations and useful outcomes. In everyday settings, the same phenomenon may happen at the early stage of machine learning tasks. Therefore, we suggest our trainable sampling method for model diversification as a viable alternative to classical methods.

\paragraph{Contributions.} \emph{Firstly}, this paper introduces Reinforced Data Sampling (RDS), a method to approximate optimum sampling for model diversification with ensemble rewarding to attain maximal machine learning potentials. A novel stochastic choice rewarding is developed as a viable mechanism for injecting model diversity in reinforcement learning.  \emph{Secondly}, we implement an end-to-end framework for trainable data sampling, which can easily be adopted in the early stages of machine learning tasks, including classification and regression.  \emph{Thirdly}, we conduct comprehensive experiments to compare RDS against other traditional data splitting methods, on real-world datasets with various tasks.  The results suggest that RDS is an effective method for data sampling with the objective of achieving high model diversification.

\section{Related Work}
\label{sec:rwork}
In machine learning, generalisation, or the ability to adapt new, previously unseen observations, plays a vital role in creating useful models \cite{cohn1994improving}. It entails the process of data sampling, which is employed to select and manipulate a representative subset of data points for performance estimation. Early approaches, such as simple random sampling or stratified sampling, have been widely adopted in numerous machine learning tasks to date. The use of simple randomisation (e.g., Knuth's algorithm \cite{knuth1997art}) in data selection is overly popular; however, it is susceptible to many sampling issues such as violation of statistical independence, bias or covariate shift \cite{may2010data,bach2010bayesian}. Stratification technique is used to partition the dataset into homogenous strata to ensure the adequate representation of data points \cite{cochran2007sampling}.

In computational learning theory, model performance and complexity have been formalised as factors to generalisation bounds according to Occam's razor \cite{rasmussen2001occam}. Hold-out method \cite{schorfheide2012use} for data selection is commonly performed to estimate the predictive performance of a learner, which can be repeated multiple times to improve stability with less variance. Furthermore, modern datasets are typically associated with heterogeneous features, ambiguous evidence, and complex dependencies, thereby leading to concept drift in model performance \cite{minku2009impact}. Importance sampling by reweighting data points has been explored as a remedial mechanism \cite{sugiyama2008direct,mahmood2014weighted}. In recent years, many researchers have approached model drift and related dataset issues with ensemble learning \cite{bach2010bayesian,minku2009impact,kolter2007dynamic,bach2008paired}. Multiple base models can be trained on blocks of data samples to address uncertainty by injecting model diversity, in the hope of maximising performance generalisation \cite{lakshminarayanan2017simple}. With recent advances in reinforcement learning \cite{peng2019trainable,buckman2018sample,lee2016stochastic}, we explore how to sample informative data points that best generalise machine learning models with ensemble learning and model diversification.

\section{Proposed Framework}
\label{sec:approach}
This study aims at developing a novel approach to sample a dataset into relevant subsets for various machine learning tasks to achieve an optimum goal. It comprises of task performance and model diversity to maximise the candidate learning potentials with adequate allocation in searching for useful models and insights in subsequent processes in machine learning. This paper formulates a sampling problem for sample-based model diversification $\delta{-div}$.

\subsection{Problem Formulation}
Let $\mathfrak{D} = \{(x_i, y_i) | x_i \in X, y_i \in Y\}_{i=1}^{M}$ denote the dataset of size $M$, where $X$ are arbitrary inputs and $Y$ are dependent outputs. We propose a data sampler 
$\omega: \{X, Y\} \to \varsigma (\{X,Y\})$ 
to generate multiple subsets $\mathfrak{d}$ of the dataset with several properties.

First, we advocate the use of diverse K learners $\mathcal{F} = \{f_1, f_2, ..., f_K\}$, including stable methods (e.g., support vector machines or regularised least square regression) and unstable methods (e.g., neural networks or decision trees). The goal is to find an optimal data sampler $\omega^*$ to maximise the ensemble learning potentials with diversity induced as the following:
\begin{equation}
  \label{eq:p1}
  \omega^*(\{X, Y\}) := \argmax_{\varsigma (\{X,Y\})}\mathfrak{E}(f_\epsilon(\varsigma (\{X,Y\}))) 
\end{equation}
where $f_\epsilon$ is an ensemble learner of $\mathcal{F}$  and $\mathfrak{E}$ is the criterion that measures the performance of $f_\epsilon$.

We posit that the sampling procedure is stochastic, in which the allocation of samples to subsets is based on parametric probability distributions $p$ with the parameters $\theta$.
\begin{equation}
  \label{eq:p2}
  \varsigma(\{x,y\}) \sim p^\theta
\end{equation}
To maintain a sampling ratio $\mathfrak{r}$, the third property of $\delta{-div}$ is described as the following:
\begin{equation}
  \label{eq:p3}
  \mu(p^\theta) \approx \mathfrak{r}
\end{equation}
where $\mu$ is the mean of the probability distributions $p^\theta$

We assume that data samples should be independent and identically distributed (i.i.d). Hence, data subsets are representative of the true population in respect to statistical independence. We formulate the fourth property for each subset $\mathfrak{d}$ of size $M_\mathfrak{d}$ as below:
\begin{equation}
  \label{eq:p4}
  (\{x_i,y_i\}) \sim \mathbb{P}(X, Y) \; \forall{i=1,...,N_\mathfrak{d}}
\end{equation}
where $\mathbb{P}$ is the true distribution.

This study is on searching for $\delta{-div}$ solutions to achieve optimum task performance; nevertheless, it is NP-Hard with the possibility of solving with approximation. Therefore, we propose a reinforcement learning approach to discovering how to sample by approximating $\delta{-div}$ solutions.

\subsection{Reinforced Data Sampling (RDS)}

We propose Reinforced Data Sampling (RDS) framework based on the Markov Decision Process (MDP) to maximise $\delta{-div}$. 

We posit the use of the data sampler $\omega$ to create a training dataset $\mathfrak{d}_{train} = \{X_{train}, Y_{train}\}$ and a test dataset $\mathfrak{d}_{test} = \{X_{test}, Y_{test}\}$ to discuss our approach without loss of generality.

RDS is a reinforcement learning framework, where an agent receives a data sample $(x, y)$ at each step, classifies which subset the sample belongs to, and interacts with an environment. As a result, a reward $r$ is given to agent by the environment based on the outcome of its action $a$. The agent reaches an optimum goal through its interactions with the environment to accumulate maximum possible rewards. It is described as a tuple $(\mathcal{S}, \mathcal{A}, \mathcal{R}, \mathcal{T})$ as the following:
\begin{itemize}
\item $\mathcal{S}$ is a finite set of states, where the decision process is evolved sample by sample.
\item $\mathcal{A}$ is the discrete action space of the agent, $\mathcal{A}(s) = \{\textit{<train>}, \textit{<test>}\}$. 
\item $\mathcal{R}$ is the reward set where $\mathcal{R}(s,a) \in \mathbb{R}$ is mapped a state $s \in \mathcal{S}$ and an action $a \in \mathcal{A}$.
\item $\mathcal{T}$ denotes the transition from the current state to the next state.
; thus, $\mathcal{T}: \mathcal{S} \times \mathcal{A} \rightarrow \mathcal{S}$.
\end{itemize}

The framework employs a stochastic policy $\pi$ which defines the probability of performing action $a$ by the agent given the state $s$; thus $\pi^\theta(a | s) \approx p^\theta(a|s)$ where the probability distribution is determined by the parameter $\theta$ according to Eq(\ref{eq:p2}). Given policy $\pi$, RDS starts from observing an initial state $s_0$ according to the probability distribution $p_0$. At each step of interaction $t$, it evolves according to:
\begin{equation}
    s_{t+1} = \mathcal{T}(s_t, a_t \sim \pi^\theta(a | s_t))
\end{equation}
We denote $\tau$ as the trajectory of the RDS, where $\tau = (s_0, a_0, s_1, a_1, ..., s_T, a_T)$. The transition $\mathcal{T}$ is deterministic as the agent moves from the current state $s_t$ to the next state $s_{t+1}$ according to the order of observations in the dataset. The optimisation problem in RDS is expressed by finding a good set of parameters $\theta$ to maximise the expected return:
\begin{equation}
    \theta^* = \argmax_{\theta}{E_\tau[R_\tau|\pi^\theta]}
\end{equation}
where $R_\tau$ is the finite-horizon undiscounted return computed based on the trajectory of $T$ steps. 

In RDS, we investigate the use of supervised learning methods, which mimic the input-output process in nature. These function approximation methods may range from linear functions to decision trees or artificial neural networks. They receive data samples as observations of the state to predict values, where $f: s \rightarrow \hat{v}(s)$. In general, the objective function for machine learning is specified as $\mathfrak{G}(s, \hat{v}(s)) \in \mathbb{R} $. At the end of each episode, we apply our function approximation on the training dataset  $\mathfrak{d}_{train} = \{ s \in \mathcal{S}| a = \textit{<train>}\}$ and the test dataset $\mathfrak{d}_{test} =\{ s \in \mathcal{S}| a = \textit{<test>}\}$. Once the transition is terminated at step $T$, we compute $R_\tau = \mathfrak{G}(\{s|a\}, \hat{v}(\{s|a\}))$.

We utilise the policy gradient method to address the optimisation problem, in which policy weights are updated by the stochastic gradient optimisation at the end of every episode as the following:
\begin{equation}
    \Delta\theta \approx \frac{\partial\log{\pi^\theta(\tau)}}{\partial\theta} R_\tau
\end{equation}
where $\pi^\theta(\tau) = \prod_{t=0}^{T}{\pi^\theta(a_t|s_t)}$ is the trajectory probability of $\tau$.

Once the policy has converged, we estimate a good approximation of our data sampler $\omega$ in Eq(\ref{eq:p1}) as the following:
\begin{equation}
    \omega(\{X, Y\})  \approx \{a_1^*, a_2^*, ..., a_T^*\}
    \label{eq:sampler}
\end{equation}
where $a_t^*= \argmax_{a \in \mathcal{A}(s_t)}{\pi^{\theta^*}(a|s_t)}$.

In general, the RDS process starts with sampling a training subset $\mathfrak{d}_{train}$ and a test subset $\mathfrak{d}_{test}$ of the dataset $\mathfrak{D}$ according to the policy $\pi$. We apply the function approximation $f$ by training on the training set $\mathfrak{d}_{train}$ and evaluating on the test set $\mathfrak{d}_{test}$ to obtain an expected return $R_\tau$. The policy $\pi$ is then updated accordingly and a new episode is started to reach the convergence of $\pi$. Refer to the Appendices \ref{app:rds} and \ref{app:algo} for details of our design and algorithm.

The convergence of RDS is inherited from the convergence of the policy gradient method with function approximation \cite{sutton2000policy}. The function approximation is designed with consistent parameter initialisation and hyper-parameters; hence, the reward is fixed for each sampled dataset. We posit that the computational complexity of RDS is $O(N(MD + CK))$, where $N$ is the number of episodes for policy updating, $M$ is the size of the dataset, $D$ is the cost for state updating, and $CK$ is the computational cost for function approximation from $K$ learners.

\subsection{Reward Mechanisms}

Our target is to train the agent to draw relevant samples with the policy $\pi$ to maximise the expected return $E$, which entails performance potentials of function approximators. In this paper, we consider generalised function approximators such as linear estimators, decision-trees, and neural networks, which are commonly useful in data challenges. Let $f$ be any arbitrary function learners, then the RDS process converges with the specified rewards \cite{sutton2000policy}.

We design the learning environment with the use of ensemble method of multiple function approximators to enrich model diversity for data selection by design; because each base model provides outcome reflecting multi-modal belief \cite{gong2019diversity}. Let $f_\epsilon$ denote the ensemble function approximator; hence we have $f_\epsilon = \Xi(f_1, f_2, ..., f_K)$, where $K$ is the number of base learning models for evaluating performance metrics. We fix the training procedure of the supervised learners, including parameter initialisation, model architecture, hyper-parameters, and random seeds to ensure the same output from a given state for reproducibility. 
This paper investigates several reward mechanisms, including soft voting and stochastic choice, which are applicable for both classification and regression problems. In the soft voting approach, we define RDS$^{DET}$ using an ensemble approximator $f^{det}_\epsilon: s \rightarrow \bar{v}(s)$ with the following value function:
\begin{equation}
    \label{eq:rds_det}
    \bar{v}(s) = \frac{1}{K} \sum_{k=1}^{K}{\hat{v_k}(s)}
\end{equation}
Thus, the environment is observed as deterministic with the reward $R_\tau = \mathfrak{G}(\{s|a\}, \bar{v}(\{s|a\}))$.

In addition, we define a stochastic RDS$^{STO}$ process $f^{sto}_\epsilon: \{f_k, k=1,2, ..., K\}$ depends on the base models randomly picked from a stationary distribution $\uprho$ at each epoch. It is desirable that this stochastic behaviour may overcome local optimisation despite the noise introduced. We define:
\begin{equation}
    \label{eq:rds_sto}
    \breve{v}(s) = \hat{v_k}(s); k \sim \uprho
\end{equation}
The environment, therefore, is observed as stochastic with the reward $R_\tau = \mathfrak{G}(\{s|a\}, \breve{v}(\{s|a\}))$.

We argue that the choice of base models is crucial to achieving higher learning potentials with model diversity. In addition, pre-processing steps or pre-trained feature mappings can be adopted in these learners to provide better representational abilities.

\subsection{Policy Optimisation}

We implement the policy learning using the Gated Recurrent Unit \cite{bahdanau2014neural} with the feature size of the dataset $\mathfrak{D}$. It is an intuitive choice as the gated networks support the data selection based on the sequence of samples similar to the desired agent's brain. In the learning, the state $s_t$ at the step $t$ is encoded to create a hidden vector presentation $h_t$ of $s_t$. With two reset and update gates, the computation of $h_{t+1}$ is described as the following:

\begin{equation}
    h_{t+1}= 
\begin{cases}
    GRU(h_t, x_{t+1} \oplus y_{t+1}) & \text{if } (x_t, y_t) \text{ is selected}\\
    GRU(h_{t-1}, x_{t+1} \oplus y_{t+1}) & \text{otherwise}
\end{cases}
\end{equation}

A linear layer is adopted to derive the probability distribution of the action $p^\theta(a|s)$. Moreover, the policy $\pi^{\theta_0}$ is pre-trained based on the sampling ratio $\mathfrak{r}$ to achieve faster convergence. 

We use the log of the action probability an equivalent loss function with the learning factor $\alpha$:
\begin{equation}
    \mathcal{L}^\theta = \alpha \frac{-\partial\log{\pi^\theta(\tau)}}{\partial\theta} R_\tau
\end{equation}

\subsubsection{Sampling Regularisation}

We implement a regularisation loss to ensure the sampling ratio based on the distributional property of the action probability in Eq(\ref{eq:p3}) as the following:
\begin{equation}
    \mathcal{L}^{r-reg} = \gamma {L1}(\mu(\pi^\theta(\tau)), \mathfrak{r})
\end{equation}
where $\gamma$ is the scale factor and $\mathfrak{r}$ is the sampling ratio.

In addition, we design a regularisation mechanism to ensure that training samples and testing samples are drawn from the same distribution as described in Eq(\ref{eq:p4}). This is important for both classification (e.g., class ratios) and regression (e.g., identically distributed). Given probability density distributions of the training set $p_{train}$ and the testing set $p_{test}$. We define:
\begin{equation}
    \mathcal{L}^{i-reg} = \psi {KL}(p_{test}, p_{train})
\end{equation}
where $\psi$ is the scale factor and ${KL}$ is the Kullback–Leibler divergence. In regression, we estimate Kullback–Leibler divergence of continuous samples using P{\'e}rez-Cruz's method \cite{perez2008kullback}.

The final loss for our policy optimisation is computed as:
\begin{equation}
    \mathcal{L} = \mathcal{L}^\theta + \mathcal{L}^{r-reg} + \mathcal{L}^{i-reg}
\end{equation}

\section{Experiments}
\label{sec:exp}

\begin{figure}[H]
    % KLP
    \centering
    \begin{subfigure}[b]{0.245\textwidth}
        \centering
        \includegraphics[width=\textwidth]{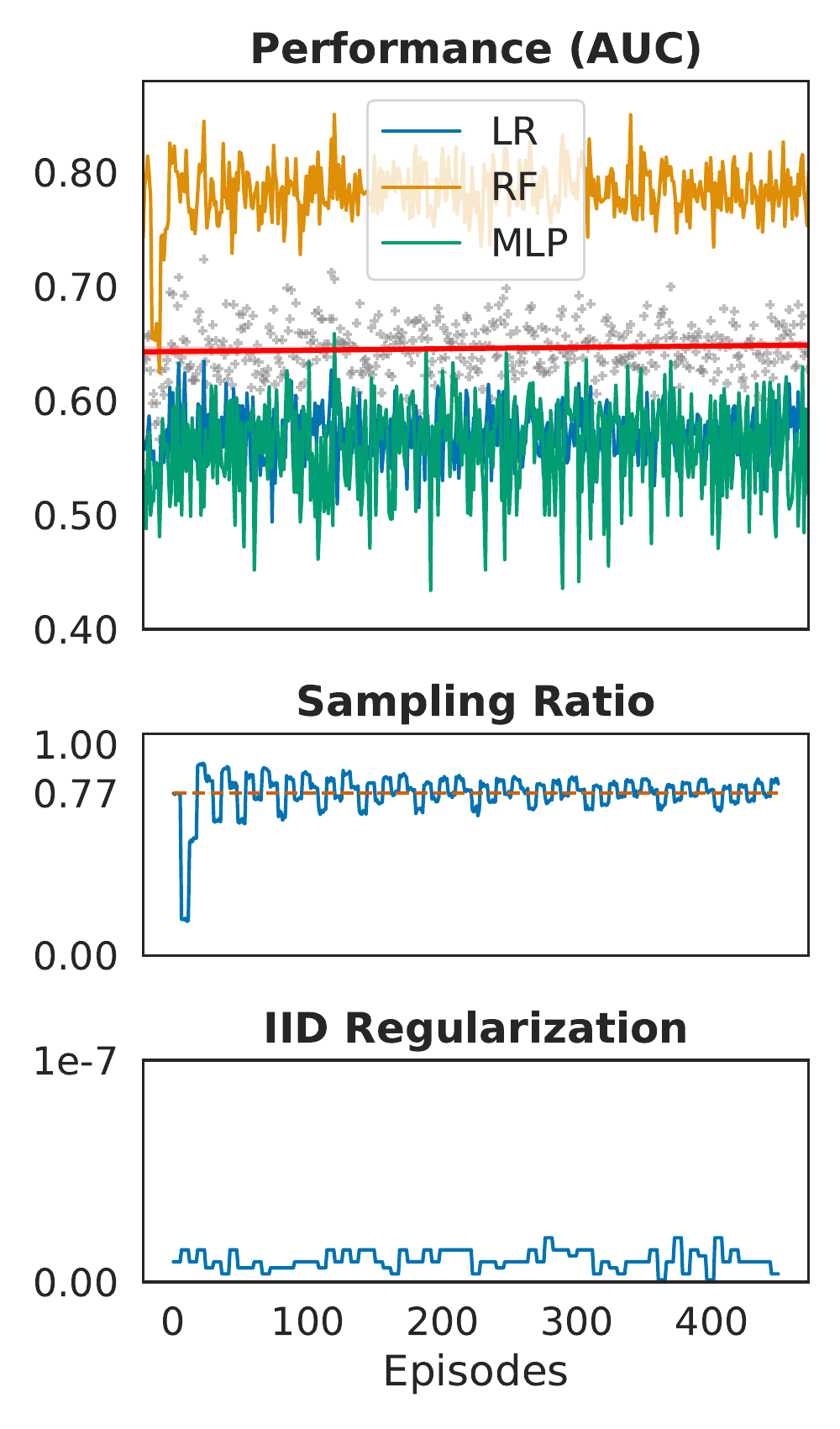}
        \caption{MADELON}
        \label{fig:mdl_det}
    \end{subfigure}
    \begin{subfigure}[b]{0.245\textwidth}
        \centering
        \includegraphics[width=\textwidth]{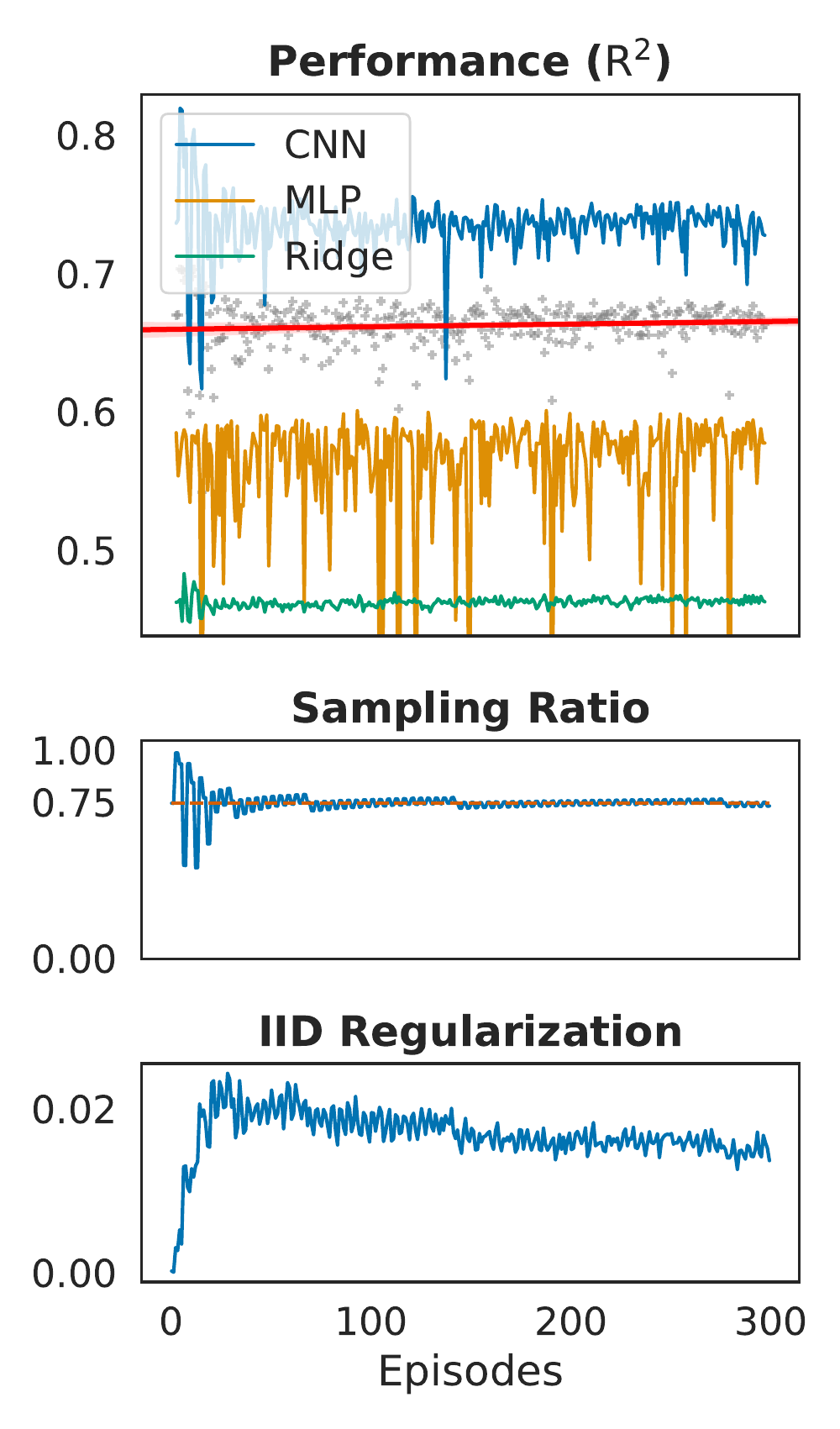}
        \caption{DR}
        \label{fig:dr_det}
    \end{subfigure}
    \begin{subfigure}[b]{0.245\textwidth}
        \centering
        \includegraphics[width=\textwidth]{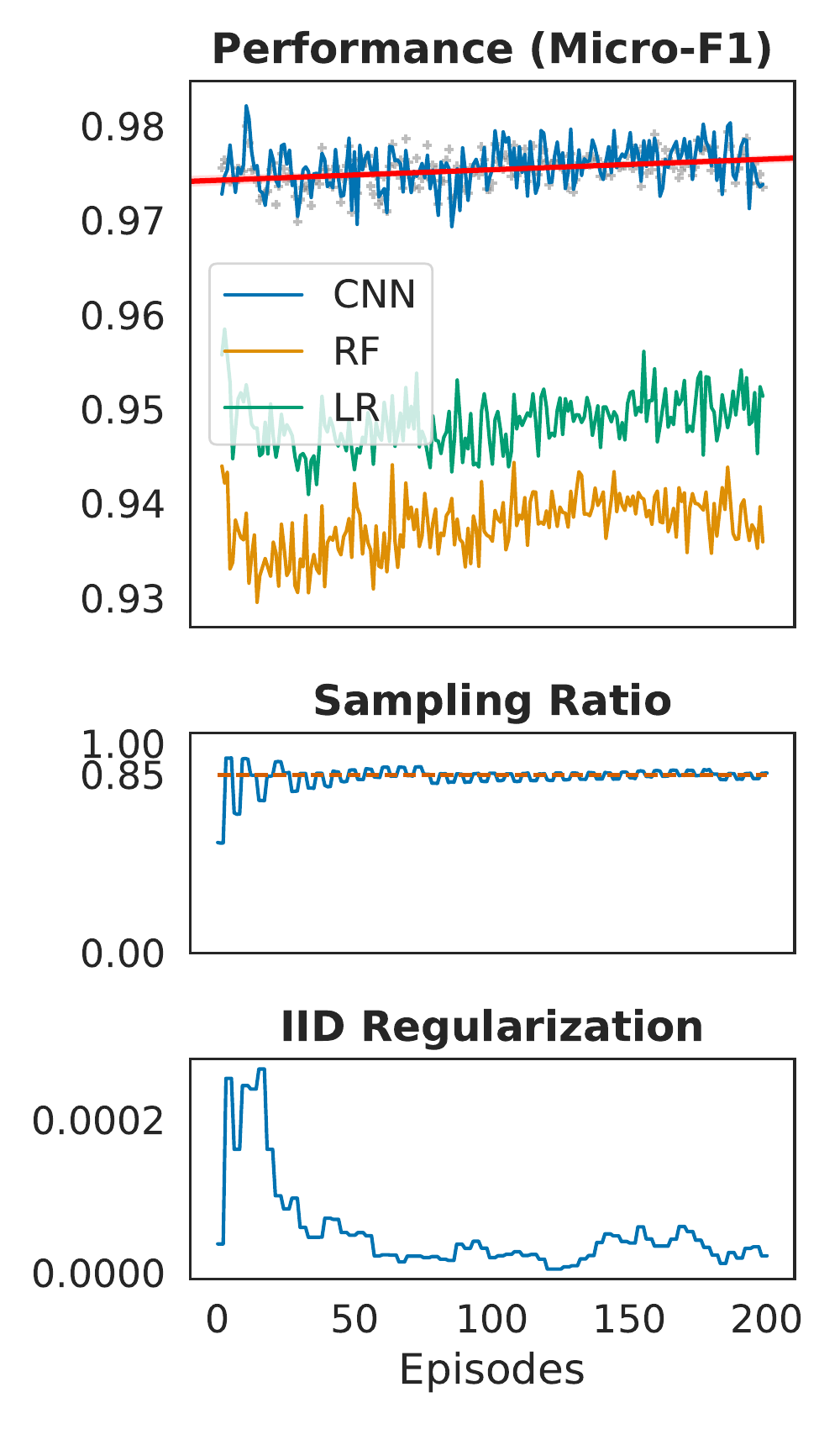}
        \caption{MNIST}
        \label{fig:mnist_det}
    \end{subfigure}
    \begin{subfigure}[b]{0.245\textwidth}
        \centering
        \includegraphics[width=\textwidth]{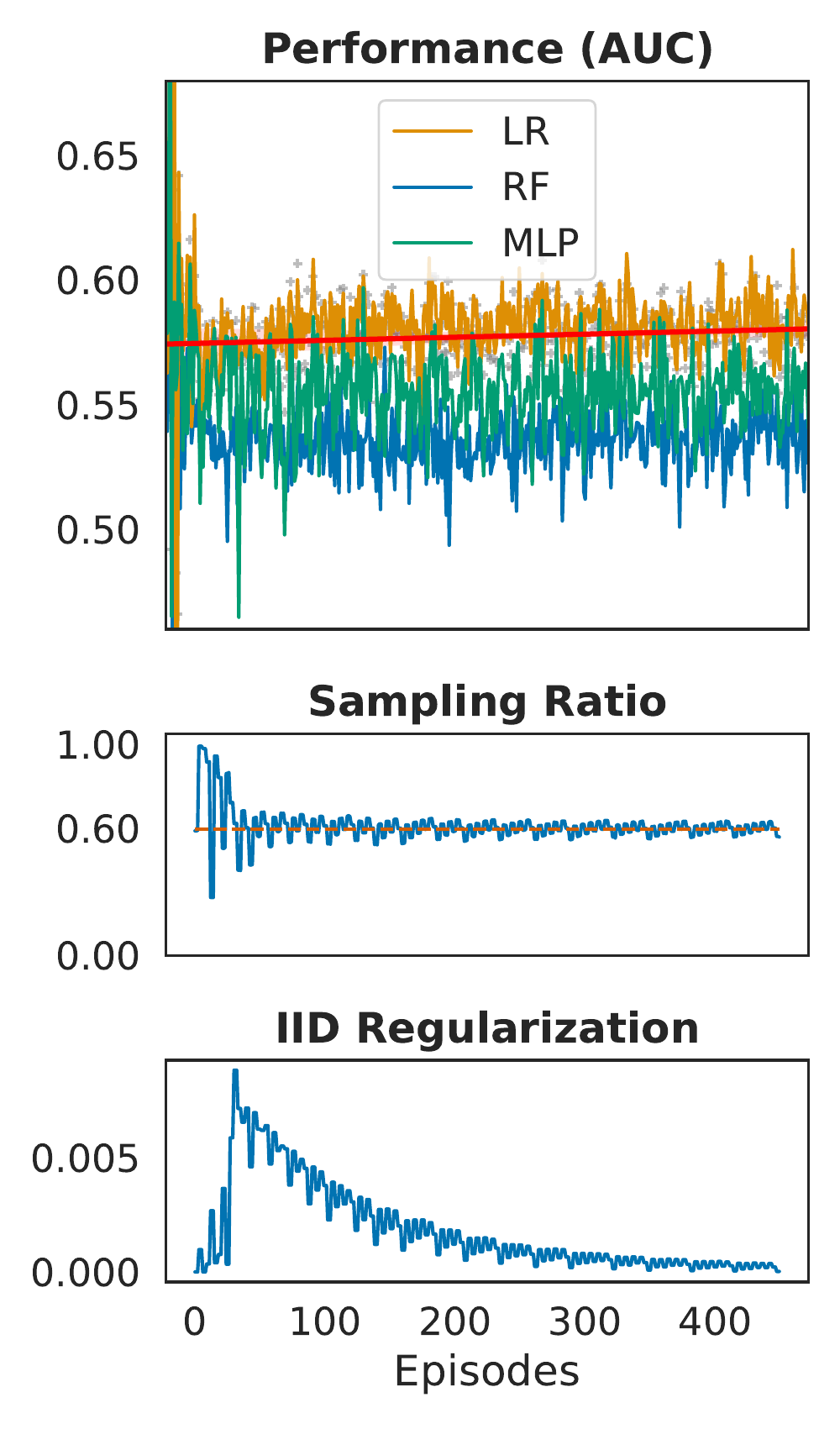}
        \caption{KLP}
        \label{fig:klp_det}
    \end{subfigure}
    \caption{Learning Dynamics for Deterministic Soft-Voting Reward Mechanism}
    \vspace{-0.2in}
    \label{fig:learning_det}
\end{figure}

\begin{figure}[H]
    % KLP
    \centering
    \begin{subfigure}[b]{0.245\textwidth}
        \centering
        \includegraphics[width=\textwidth]{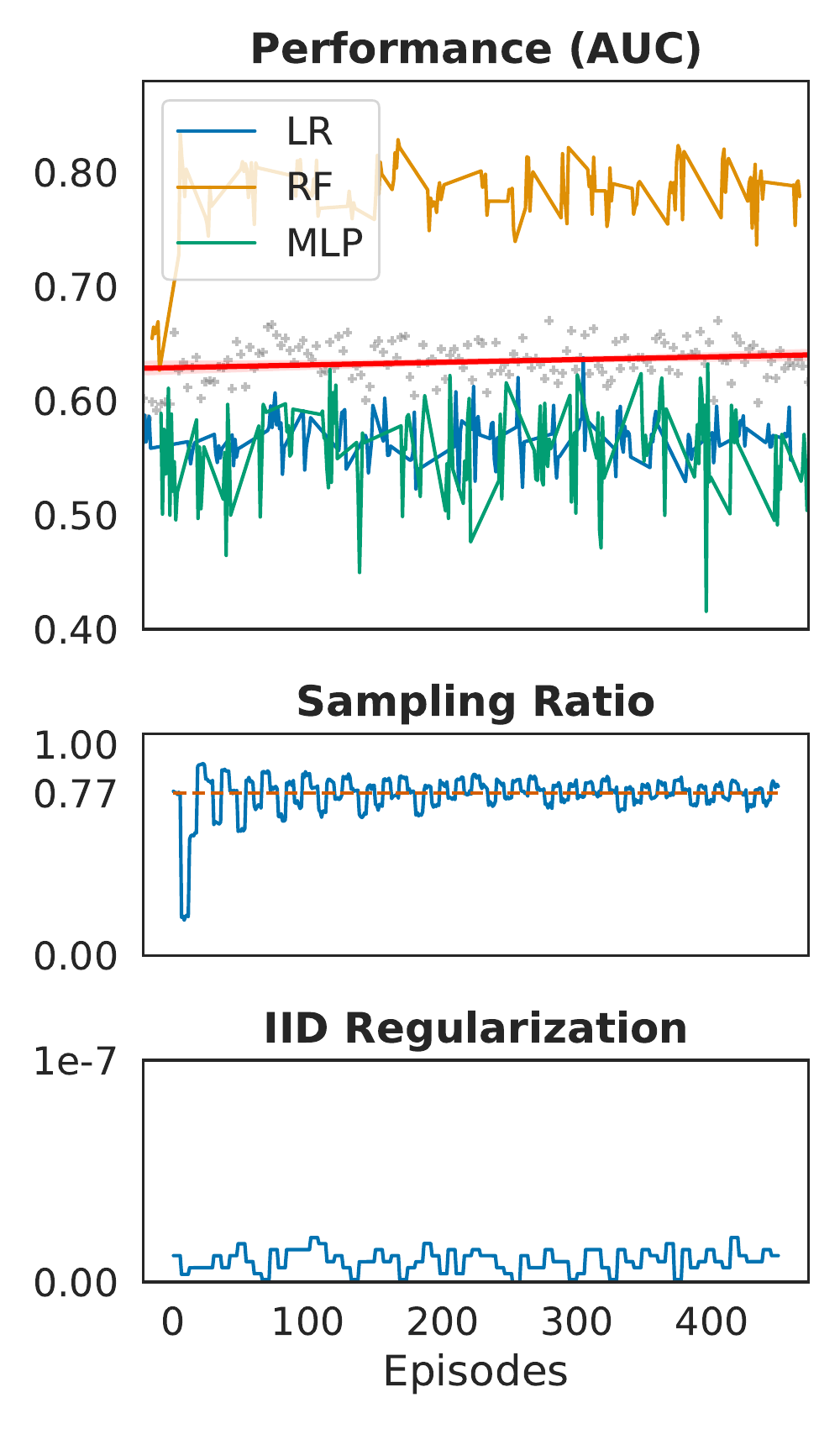}
        \caption{MADELON}
        \label{fig:mdl_sto}
    \end{subfigure}
    \begin{subfigure}[b]{0.245\textwidth}
        \centering
        \includegraphics[width=\textwidth]{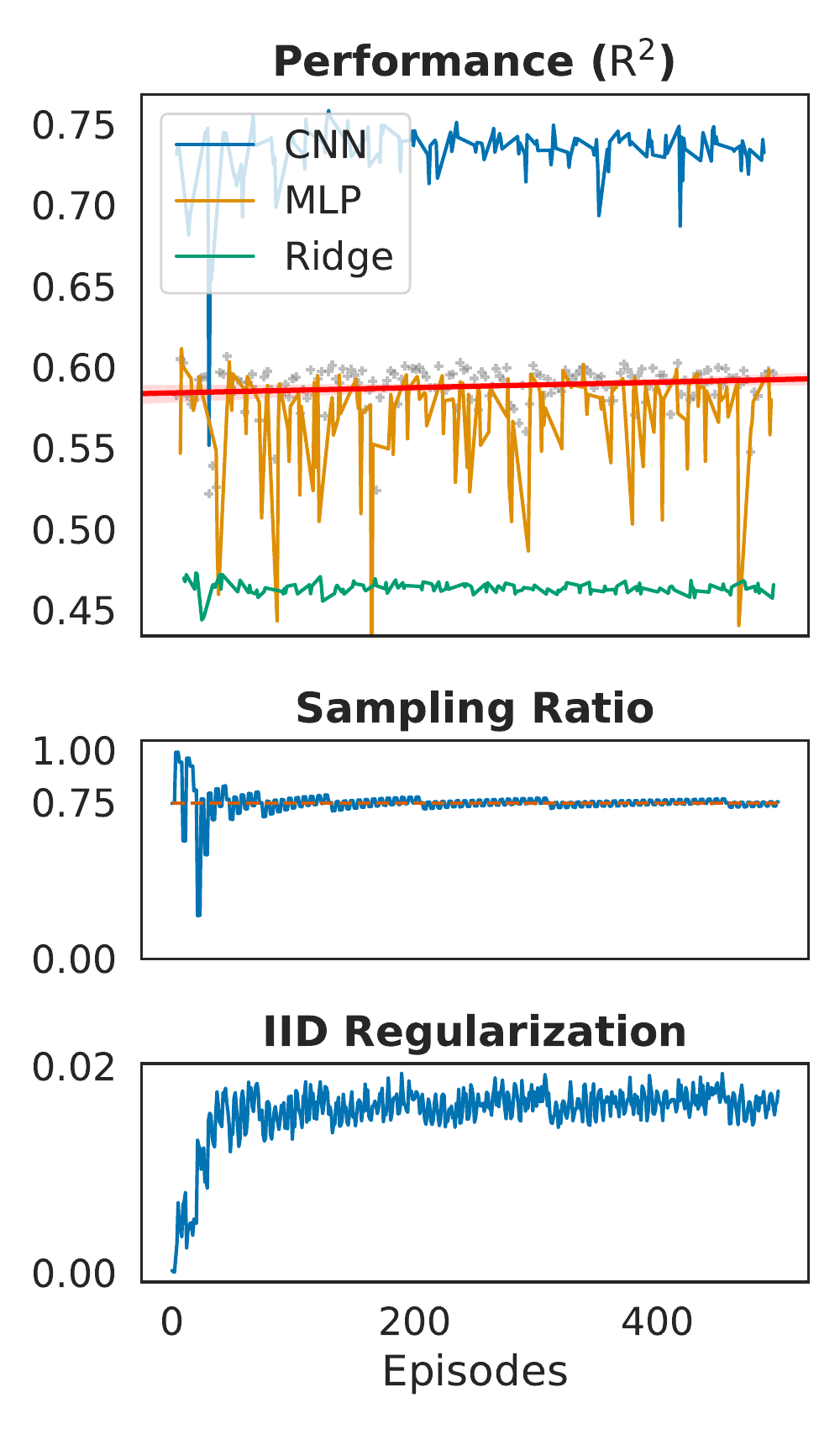}
        \caption{DR}
        \label{fig:dr_sto}
    \end{subfigure}
    \begin{subfigure}[b]{0.245\textwidth}
        \centering
        \includegraphics[width=\textwidth]{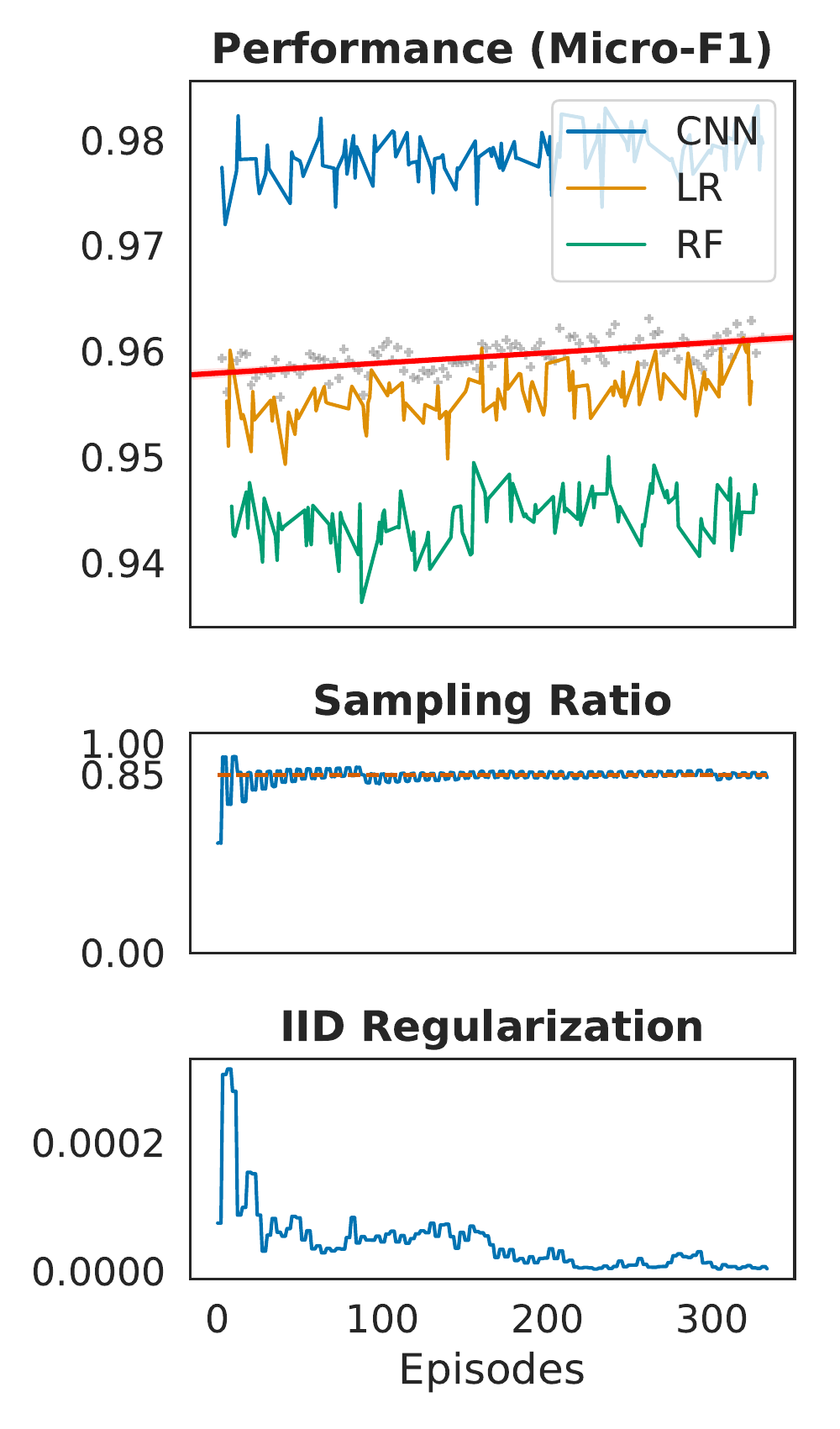}
        \caption{MNIST}
        \label{fig:mnist_sto}
    \end{subfigure}
    \begin{subfigure}[b]{0.245\textwidth}
        \centering
        \includegraphics[width=\textwidth]{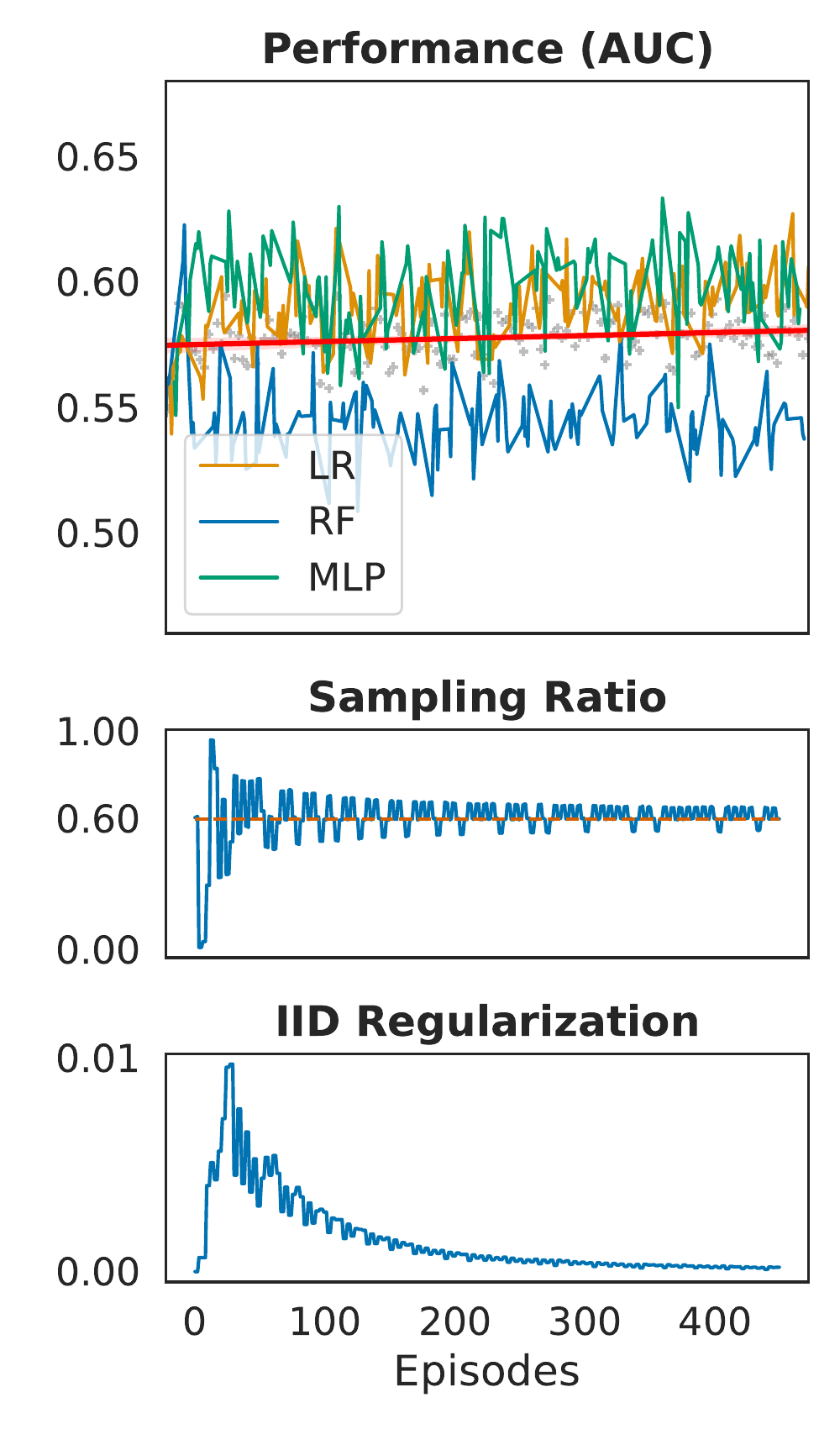}
        \caption{KLP}
        \label{fig:klp_sto}
    \end{subfigure}
    \caption{Learning Dynamics for Stochastic Choice Reward Mechanism.}
    \vspace{-0.2in}
    \label{fig:learning_sto}
\end{figure}

In this section, we conduct experiments on four datasets to examine the effectiveness of the RDS method. It is done via evaluating model diversification reflected by the performance evaluated on proposed data samples by RDS in comparison with classical methods.

\textbf{Madelon (MDL)}~\cite{guyon2005result} was artificially developed for the NIPS 2003 feature selection challenge. It has 500 numerical features, in which 20 real features and 480 distractors have no predictive capacity. Several pre-processing techniques were adopted to conceal the origin and patterns of the dataset on the search for functional feature extractors. We employ bare-bone Logistic Regression (LR), Random Forest (RF), and Multi-Layer Perception (MLP) in our experiments. And a pipeline of stability selection and Logistic Regression of feature interactions is adopted for public benchmarking \cite{cannon2018madelon}.

\textbf{Drug Review (DR)}~\cite{grasser2018aspect} provides patients' reviews on specific drugs crawled from multiple online pharmaceutical review sites.  It contains categorical features including drug name and patient condition, review text and date, and numerical features including review rating and useful counts.  In total, there are 215,063 examples, which is split into a training set of 75\% and a test set of 25\%.   In this experiment, we use three base learners including Ridge Regression (Ridge), Multi-Layer Perception (MLP), and Convolutional Neural Network (CNN).

\textbf{MNIST}~\cite{lecun1998mnist} consists of 70,000 hand-written digits and is one of the most well-known datasets in the deep learning community. MNIST is selected for experiments since it represents very well for multi-class classification task on images. MNIST considered a balanced image classification data, and it is divided into 60K samples for training and 10K samples for testing. 
In this experiment, we use three base learners including Logistic Regression (LR), Random Forest (RF), and Convolutional Neural Network (CNN).

\textbf{Kalapa Credit Scoring (KLP)}~\cite{klp2020credit} is a data challenge for credit scoring task. The dataset consists of 30,000 training and 20,000 testing examples. It contains two labels (i.e., GOOD and BAD) associated with 62 variables, including demographics and financial status. There is an imbalance problem on the label distribution with a ratio of 1.6\% (i.e., only 486 BAD samples among 30,000 training samples). Moreover, 40 data fields have missing rates of more than 30\%, which increases the difficulty in finding a good data selection for the data challenge. We will consider three models namely Logistic Regression (LR), Random Forest (RF), and Multi-Layer Perception (MLP) on investigating the effectiveness of the splitting methods. The first ranked solution \cite{klp2020sol} is selected as the public model for comparisons.

\begin{table}[t]
\caption{Madelon Experiment. \emph{Public} denotes a public solution of~\cite{cannon2018madelon}.}
\label{tab:mdl_results}
\centering
\small
\begin{tabular}{lrrrrrrrcc}
\toprule
\multirow{2}{*}{Sampling} & \multicolumn{2}{c}{\#Sample} & \multicolumn{2}{c}{Class Ratio} & \multirow{2}{*}{LR} & \multirow{2}{*}{RF} & \multirow{2}{*}{MLP} & \multirow{2}{*}{Ensemble} & \multirow{2}{*}{Public} \\
 & Train & Test & Train & Test &  &  &  &  &  \\
\midrule
Preset & 2000 & 600 & 1.0000 & 1.0000 & .6019 & \textbf{.8106} & .5590 & .6783 & .9063 \\
Random & 2000 & 600 & .9920 & 1.0270 & .5742 & .7729 & .5774 & .6453 & .9002 \\
Stratified & 2000 & 600 & 1.0000 & 1.0000 & .5673 & .7470 & .6153 & .6360 & .8828 \\
RDS$^{DET}$ & 2001 & 599 & 1.0375 & .9137 & \textbf{.6192} & .8050 & \textbf{.6228} & \textbf{.6973} & .8915 \\
RDS$^{STO}$ & 2021 & 579 & 1.0010 & .9966 & \textbf{.6192} & .8050 & .6050 & .6947 & \textbf{.9106} \\
\bottomrule
\end{tabular}
\end{table}

\begin{table}[t]
\caption{Drug Review Experiment. \emph{Public} denotes a public solution, Bi-LSTM with Attention, on Kaggle.}
\label{tab:dr_results}
\centering
\small
\begin{tabular}{lccccccc}
\toprule
Sampling & Train & Test & Ridge & MLP & CNN & Ensemble & Public \\
\midrule
Preset & 161,297 & 53,766 & .4580 & \textbf{.5787} & .7282 & .6660 & .7637 \\
Random & 161,297 & 53,766 & .4597 & .4179 & .7353 & .6485 & .7503 \\
RDS$^{DET}$ & 162,070 & 52,993 & .4646 & .5776 & .7355 & \textbf{.6692} & \textbf{.7649} \\
RDS$^{STO}$ & 161,944 & 53,119 & \textbf{.4647} & .5370 & \textbf{.7509} & .6562 & .7600 \\
\bottomrule
\end{tabular}
\end{table}

\subsection{Experimental Settings}

\paragraph{Implementation.} The source code of RDS is implemented in Pytorch whilst learning environments are built flexibly by various learning frameworks such as Keras, Tensorflow or Scikit-learn. Environmental learning models are optimised concurrently using a common evaluation metric. For the policy optimisation, the number of hidden unit of GRU is 256. The learning is run on 3-400 episodes with the RMSprop optimiser and the initial learning rate of 0.001. Scaling factors $(\alpha, \gamma, \omega)$ are empirically selected, i.e., (1.0, 0.9, 0.1) for Madelon, (1.0, 1.0, 40) for Kalapa, (1.0, 0.1, 0.01) for MNIST and (1.0, 0.9, 0.1) for Drug Review. For KLP and DR datasets, we employ FastText~\cite{bojanowski2016enriching} and BERT~\cite{devlin-etal-2019-bert} language models for extracting representation for textual contents. All experiments are conducted on a similar computational environment of Intel(R) Xeon(R) Gold 6244 CPU @ 3.60GHz, 256GB Ram, and a Titan RTX 2080Ti GPU card. 

\paragraph{Baselines.} We compare our proposed RDS approach with several traditional data sampling methods, including simple randomisation denoted as \emph{Random}, stratification (only for classification) denoted as \emph{Stratified}. 
We also include comparisons with available splitting denoted as \emph{Preset}, which is provided either by the organisers of competitions or the authors of the datasets. Moreover, we select a number of prominent solutions which have been shared by the members of the public to examine the effects of various techniques on the datasets.

\paragraph{Metrics.} For Madelon and Kalapa datasets, the tasks are binary classification; therefore, we use AUC to measure model performance.  In turn, we employ Micro-F1 metric for the task of multi-class classification on MNIST dataset.  For experiment on Drug Review, we use R-squared ($\mathrm{R}^2$) to measure performance of the models as the task is regression.

\subsection{Results and Discussion}

\begin{table}[t]
\caption{MNIST Experiment. \emph{Public} is a solution based on CNN architecture. It is noted that the public solution is not the same CNN's architecture used in RDS, which has fewer layers and takes shorter time to train.}
\label{tab:mnist_results}
\centering
\small
\begin{tabular}{lccccccccc}
\toprule
\multirow{2}{*}{Sampling} & \multicolumn{2}{c}{\#Sample} & \multicolumn{2}{c}{Class Ratio} & \multirow{2}{*}{LR} & \multirow{2}{*}{RF} & \multirow{2}{*}{CNN} & \multirow{2}{*}{Ensemble} & \multirow{2}{*}{Public} \\
 & Train & Test & Train & Test &  &  &  &  &  \\
\midrule
Preset   & 60000 & 10000 & .8571 & .1429 & .9647 & \textbf{.9524} & .9824 & .9819 & .9917 \\
Random   & 59500 & 10500 & .8500   & .1500 & .9603 & .9465 & .9779 & .9768 & .9914 \\
Stratified & 59500 & 10500 & .8500  & .1500 & \textbf{.9625} & .9510  & .9795 & .9792 & .9901 \\
RDS$^{DET}$ & 59938 & 10062 & .8562 & .1438  & .9495 & .9382 & .9757 & .9769 & .9927 \\
RDS$^{STO}$ & 59496 & 10504 & .8499 & .1501 & .9583 & .9486 & \textbf{.9851} & \textbf{.9830} & \textbf{.9931} \\
\bottomrule
\end{tabular}
\end{table}

\begin{table}[t]
\caption{Kalapa Experiment. \emph{Public} denotes the first solution from Kalapa challenge on the private leader board.}
\label{tab:klp_results}
\centering
\small
\begin{tabular}{lccccccccc}
\toprule
\multirow{2}{*}{Sampling} & \multicolumn{2}{c}{\#Sample} & \multicolumn{2}{c}{Class Ratio} & \multirow{2}{*}{LR} & \multirow{2}{*}{RF} & \multirow{2}{*}{MLP} & \multirow{2}{*}{Ensemble} & \multirow{2}{*}{Public} \\
 & Train & Test & Train & Test &  &  &  &  &  \\
\midrule
Preset & 30000 & 20000 & .0165 & .0186 & .5799 & .5517 & .5635 & .5723 & .5953 \\
Random & 30000 & 20000 & .0169 & .0179 & .5886 & .5374 & .5914 & .5856 & .6042 \\
Stratified & 30000 & 20000 & .0173 & .0173 & .5952 & \textbf{.5608} & .5780 & .5983 & .6014 \\
RDS$^{DET}$ & 29999 & 20001 & .0180 & .0163 & \textbf{.6045} & .5350 & .5802 & .6057 & .5362 \\
RDS$^{STO}$ & 30031 & 19969 & .0172 & .0174 & .5997 & .5491 & \textbf{.6354} & \textbf{.6072} & \textbf{.6096}\\
\bottomrule
\end{tabular}
\end{table}

The results demonstrate that our proposed RDS approach with various reward mechanisms works steadily with the four datasets. \autoref{fig:learning_det} depicts the learning dynamics of RDS$^{DET}$, in which the regression line is highlighted in red to indicate the improvements over time of the designed agent with diversification of multiple base models. Given a finite number of episodes, RDS$^{DET}$ establishes desirable optimisation behaviours regularised by sampling assumptions of the $\delta{-div}$ problem. Likewise, \autoref{fig:learning_sto} illustrates the learning dynamics of stochastic reward mechanism, in which lesser numbers of approximations are exhibited in the model performance of all datasets. The results show better optimisation of the learning gradients with this simple yet efficient method.

In details, RDS$^{DET}$ yields good performance for the ensemble performance which has been directly optimised for. This upward trend can be clearly observed across all datasets. RDS$^{STO}$ demonstrates clear outperformance for the base learners, especially the results are significant for LR model on \emph{Madelon} (Table~\ref{tab:mdl_results}), CNN models on \emph{DR} (Table~\ref{tab:dr_results}) and \emph{MNIST} (Table~\ref{tab:mnist_results}), as well as \emph{KLP} (Table~\ref{tab:klp_results}). Amongst the baselines, \emph{Stratified} has a strength of maintaining class ratios for the task of classification, which can also be maintained by the proposed RDS methods. The \emph{Preset} splitting, given by the competition organiser or authors, appears to be either \emph{Random} or \emph{Stratified}. Thus, they obtain comparable performance to randomisation and stratification but worse than RDS variants. Although the preset allocation performs well in some settings, the adequate performance of the RDS is consistently observed in both ensemble evaluation and public benchmarking. The stochastic choice mechanism gains some advantages over the previously designed algorithms. Moreover, the assumption of statistical independence holds a critical impact on the learning of the agent, which must be considerably regularised for imbalance datasets. See Appendix \ref{app:notes} for experiment notes.

Trainable data sampling for model diversification achieves good performance based on ensemble learning and publicly available solutions; thus, higher learning potentials are yet to be explored.

\section{Conclusion}
This paper proposes Reinforced Data Sampling (RDS) method, which learns to select representative samples. The objective is to emphasise model diversification $\delta{-div}$ by maximising learning potentials of various base learners. We introduce different reward mechanisms including soft voting and stochastic choice to train optimal sampling policy under reinforcement learning framework. Experiments conducted on four datasets evidently highlight the benefits of using RDS over classical sampling approaches. Moreover, RDS's sampling approach is configurable and can be applied to many different types of data and models. 

\clearpage

\section{Broader Impact}
This research is one fundamental step in advances of information processing that may levitate many tasks in machine learning. Our proposed Reinforced Data Sampling (RDS) approach will bring meaningful changes to the research community and related industries. In practice, we advocate that the use of RDS is preferable over the popular selection methods, such as simple randomisation, stratification, or hold-out, in classification and regression. Promoting optimum sampling with model diversity will also bring far-reaching impacts in the hope of searching for useful models and insights in diverse venues, including worldwide AI challenges and large-scale research projects. During our research, we have contacted multiple participants and winners of recent data challenges with sizable monetary prizes; and improper data selection with concept drift was the key issue causing the waste of vast amounts of hours and computational powers. In average, each AI competition yields around hundreds to thousands of individuals or teams with enormous resources. The adoption of our framework, therefore, has potential environmental impacts to minimise the possible loss of excessive experimentation and productivity. In addition, model diversification will also be beneficial for researchers, competition organisers and large companies to reach maximal potentials of data and models. 

\bibliographystyle{unsrt}
\bibliography{document}

\newpage
\appendix
\section{Appendices}
This section covers the supplementary information for our approach and experiments as the following:
\begin{itemize}
\setlength{\itemsep}{0pt}
\setlength{\parskip}{0pt}
\setlength{\parsep}{0pt}
\item Overall Process of Reinforced Data Sampling for Model Diversification
\item RDS Algorithms
\item Datasets
\item Detailed Experiment Specifications and Notes
\end{itemize}

\subsection{Overall RDS Process}
\label{app:rds}
The RDS process starts with the initialisation of the sampling policy with an initial distribution $p_0$ based on the sampling ratio $\mathfrak{r}$ in Eq(\ref{eq:p3}). During an episodic run, an action $a$ is drawn from the policy $\pi^\theta$ for each state $s$ based on the trajectory of $T$ steps. The environment employs base learners to handle data samples $(x, y)$ based on the given action $a$ and transits to the next step. RDS stores the approximated values $v$ of the data samples into a replay memory; which is used to compute the ensemble return $R_\tau$ for policy update. %RDS$^{DET}$ uses soft-voting mechanism to compute $\bar{v}$ in Eq(\ref{eq:rds_det}); whereas, RDS$^{STO}$ makes stochastic choice to compute $\breve{v}$ as shown Eq(\ref{eq:rds_sto}). \autoref{fig:rds} provides a summary of the reinforced sampling process.

\begin{figure}[h!]
    \centering
    \includegraphics[width=0.75\textwidth]{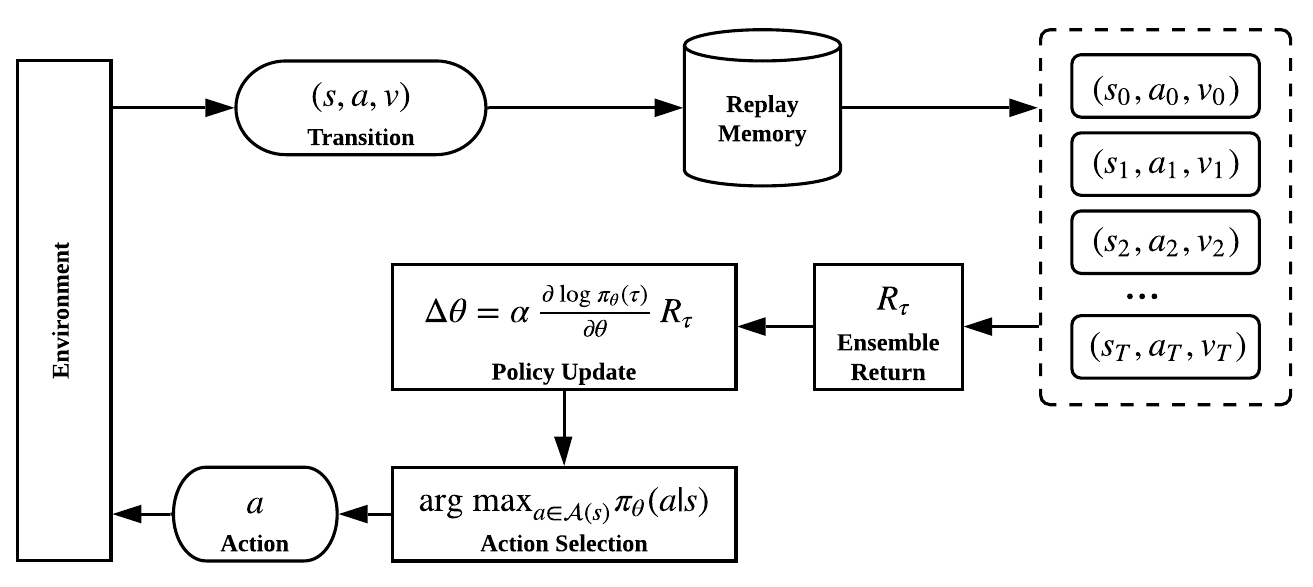}
    \caption{Overall Process of Reinforced Data Sampling (RDS)}
    \vspace{-0.2in}
    \label{fig:rds}
\end{figure}

\subsection{Algorithms}
\label{app:algo}
This subsection describes the algorithms for two variants of the RDS as the following.

{\centering
\begin{minipage}{.8\linewidth}
\begin{algorithm}[H]
\caption{Training Algorithm for RDS$^{DET}$}\label{alg:training_det}
\begin{algorithmic}[1]
\Require dataset $\mathfrak{D} = \{(x_t, y_t) | x_t \in X, y_t \in Y\}_{t=1}^{T}$, ensemble learner $f_\epsilon$, sampling ratio $\mathfrak{r}$
\Procedure{RDS}{$\mathfrak{D},f_\epsilon,\mathfrak{r}$}
\State \textbf{initialise} $\pi^\theta \gets pretrain(\mathfrak{r})$
\ForAll{$i \gets 1, L$} \Comment{$L$ is the maximum number of epochs}
\State $V \gets \emptyset$ \Comment{$V$ is the replay memory}
\For{$t \gets 1,T$}
\State $s_t \gets (x_t, y_t)$
\State \textbf{select} action $a_t \sim \pi^\theta(a|s_t)$
\State $V \gets V \cup \{(s_t, a_t)\}$ \Comment{Saving to the replay memory}
\EndFor
\State $\mathfrak{d}_{train} \gets \{ s \in V | a = \textit{<train>}\}$
\State \textbf{train} $f^{DET}_\epsilon(\cdot)$ on the sampled training set $\mathfrak{d}_{train}$
\State $\mathfrak{d}_{test} \gets \{ s \in V | a = \textit{<test>}\}$
\State \textbf{compute} $R_\tau = \mathfrak{G}(\mathfrak{d}_{test}, f^{DET}_\epsilon(\mathfrak{d}_{test}))$ \Comment{According to Eq(\ref{eq:rds_det})}
\State \textbf{update} the policy $\pi^\theta$
\EndFor
\State \textbf{generate} $\omega(\{X, Y\})$ \Comment{According to Eq(\ref{eq:sampler})}
\State $\mathfrak{d}_{train}, \mathfrak{d}_{test} \gets \omega(\{X, Y\})$
\State \textbf{return} $(\mathfrak{d}_{train}, \mathfrak{d}_{test})$
\EndProcedure
\end{algorithmic}
\end{algorithm}
\end{minipage}
\par
}

{\centering
\begin{minipage}{.8\linewidth}
\begin{algorithm}[H]
\caption{Training Algorithm for RDS$^{STO}$}\label{alg:training_sto}
\begin{algorithmic}[1]
\Require dataset $\mathfrak{D} = \{(x_t, y_t) | x_t \in X, y_t \in Y\}_{t=1}^{T}$, ensemble learner $f_\epsilon$, sampling ratio $\mathfrak{r}$
\Procedure{RDS}{$\mathfrak{D},f_\epsilon,\mathfrak{r}$}
\State \textbf{initialise} $\pi^\theta \gets pretrain(\mathfrak{r})$
\ForAll{$i \gets 1, L$} \Comment{$L$ is the maximum number of epochs}
\State $V \gets \emptyset$ \Comment{$V$ is the replay memory}
\For{$t \gets 1,T$}
\State $s_t \gets (x_t, y_t)$
\State \textbf{select} action $a_t \sim \pi^\theta(a|s_t)$
\State $V \gets V \cup \{(s_t, a_t)\}$ \Comment{Saving to the replay memory}
\EndFor
\State $\mathfrak{d}_{train} \gets \{ s \in V | a = \textit{<train>}\}$
\State \textbf{choose} $f^{STO}_\epsilon(\cdot)$ from the stationary distribution $\uprho$
\State \textbf{train} $f^{STO}_\epsilon(\cdot)$ on the sampled training set $\mathfrak{d}_{train}$
\State $\mathfrak{d}_{test} \gets \{ s \in V | a = \textit{<test>}\}$
\State \textbf{compute} $R_\tau = \mathfrak{G}(\mathfrak{d}_{test}, f^{STO}_\epsilon(\mathfrak{d}_{test}))$ \Comment{According to Eq(\ref{eq:rds_sto})}
\State \textbf{update} the policy $\pi^\theta$
\EndFor
\State \textbf{generate} $\omega(\{X, Y\})$ \Comment{According to Eq(\ref{eq:sampler})}
\State $\mathfrak{d}_{train}, \mathfrak{d}_{test} \gets \omega(\{X, Y\})$
\State \textbf{return} $(\mathfrak{d}_{train}, \mathfrak{d}_{test})$
\EndProcedure
\end{algorithmic}
\end{algorithm}
\end{minipage}
\par
}

\subsection{Datasets}

In this paper, four datasets are selected for the analysis of the effectiveness of RDS. They cover a range of machine learning tasks, including binary classification, multi-class classification and regression. Our study simulates the way that the datasets are prepared for the tasks, given no prior knowledge on existing or public solutions. 

We employ bare-bone base models with minimal settings to conduct experiments on the datasets. They are mostly originated from data science or AI challenges. The existing train/test subsets of the datasets are obtained from their public websites, forums, or emails.

The primary purpose is to demonstrate the generalisability of RDS by applying base models for data sampling and evaluating the data samples with ensemble learning and publicly available solutions. We note that the existing solutions may have been designed to fit the preset allocation of the datasets. For example, hyper-parameters or hand-craft rules may have been explicitly fine-tuned to the published data samples to perform highly in the competitions. Nevertheless, it is intriguing to examine the effects of RDS using existing public solutions, knowing the learning potentials of trainable data samples may even go further.

\begin{table}[H]
\caption{List of datasets used in this paper.}
\label{tab:datasets}
\resizebox{\textwidth}{!}{
\centering
\begin{tabular}{lllllrrcr} 
\toprule
Dataset & Task & Challenge & Size of Data & Evaluation & Year \\ 
\midrule
MADELON & Binary Classification & NIPS 2013 Feature Selection  & $2,600 \times 500$ (multivariate)  & AUC & 2003 \\
DR & Regression & Drug Reviews (Kaggle Hackathon) & $215,063 \times 6$ (multivariate, text) & $R^2$ & 2018 \\
MNIST & Multiclass Classification & Hand Written Digit Recognition &  $70,000 \times 28 \times 28$ (image) & Micro-F1 & 1998 \\
KLP & Binary Classification & Kalapa Credit Scoring & $50,000 \times 64$ (multivariate, text) & AUC & 2020 \\
\bottomrule
\end{tabular}
}
\vspace{0.5in}
\end{table}

\subsection{Experiment Specifications and Notes}
\label{app:notes}
This subsection describes the specifications and notes of our experiments in great details.

\newpage
\begin{table}[H]
  \centering
    \begin{tabularx}{\textwidth}{lX}
    \toprule
    Experiment     & Notes \\
    \midrule
    Background & Madelon was artificially developed for the NIPS 2003 feature selection challenge. It has 500 numerical features, in which 20 real features and 480 distractors have no predictive capacity. Several pre-processing techniques were previously adopted to conceal the origin and patterns of the dataset on the search for functional feature extractors. \\
    Settings & (1.0, 0.9, 0.1) \\
    Base Models & \tabitem \textbf{LR} - Logistic Regression (solver='liblinear',penalty='l2',random\_state=123)  \\
                & \tabitem \textbf{RF} - Random Forest(n\_estimators=128,random\_state=123) \\
                & \tabitem \textbf{MLP} - Multilayer Perceptron (Adam,lr=1e-3,manual\_seed=123) \\
                & \tabitem \textbf{SVC} - Support Vector Classifier (kernel='rbf',coef0=1) \\
    Benchmark & A pipeline of stability selection and Logistic Regression of feature interactions is adopted for public benchmarking \cite{cannon2018madelon}\\
    Pre-processing & No pre-processing needed \\
    Run time & 90s / epoch \\
    Observations & \\
    & \textbf{Learning Dynamics} as shown in \autoref{fig:learning_det}(a) and \autoref{fig:learning_sto}(a) \\
    & \tabitem LR has performed consistently on the Madelon dataset. There were some low performance points at the beginning; however, the upward trend can be observed during the agent learning. \\
    & \tabitem Both LR and MLP are less stable with lower performance than RF. \\
    & \tabitem The ensemble performance falls between the range of the performance of three base models. This observation indicates that there are disagreements among classifiers, which hints that model diversity is injected into model performance. \\
    & \tabitem The regularisations serve as important mechanisms for optimum allocation. \\
    & \tabitem There is the balance in class ratios of the dataset; hence, IID regularisation is approaching almost zero. \\
    & \tabitem Experiments on SVM yield similar observations at higher computational cost; thus, SVM results are not included in our report. \\
    & \textbf{Comparison between RDS$^{DET}$ and RDS$^{STO}$} \\
	& \tabitem Consistent results are observed on both deterministic and stochastic reward mechanisms. \\
	& \tabitem The agent learning has become stable within the first 30 episodes. \\
	& \tabitem The stochastic choice has the same learning dynamics with the noticeable, lesser number of value approximations. \\
	& \textbf{Comparison between RDS and other methods as shown in \autoref{tab:mdl_results}} \\
	& \tabitem The preset selection shows the highest performance on RF; which may hint that the dataset was prepared with holdout based on RF. However, it has the worst performance on MLP; therefore, it may hinder neural network-based solutions during the challenge. \\
	& \tabitem Both random and stratified sampling techniques show sub-optimal performance in all evaluation metrics.  \\
	& \tabitem RDS has a good balance amongst all three base models. The performance of RDS on RF is slightly lower than the performance of the preset selection. However, it yields higher results in both LR, MLP and Ensemble. \\
	& \tabitem RDS$^{DET}$ has achieved the highest performance in the ensemble use of three base models. \\
	& \tabitem Preset and stratified allocation have a perfect sampling ratio. \\	
	& \textbf{Public Benchmarking} \\
	& \tabitem The use of public solution has shown higher performance compared to the bare-bone ensemble of three base models. The evaluation metrics are in agreement across multiple sampling techniques, except RDS$^{DET}$. \\
	& \tabitem The RDS$^{STO}$ performs as the top in public benchmarking. \\
    & \textbf{Summary} \\
	& \tabitem The Madelon experiment highlight effectiveness of reinforced sampling for model diversification. \\
	& \tabitem The model diversity is observed based on the ensemble and performance of various base models. \\
	& \tabitem The regularisations play an important role in optimum sampling. \\
    \bottomrule
    \end{tabularx}%
  \label{tab:madelon_notes}%
  \caption{Madelon - Experiment Notes}
\end{table}

\begin{table}[H]
  \centering
    \begin{tabularx}{\textwidth}{lX}
    \toprule
    Experiment     & Notes \\
    \midrule
    Background & Drug Review (DR)~\cite{grasser2018aspect} provides patients' reviews on specific drugs crawled from multiple online pharmaceutical review sites.  It contains categorical features including drug name and patient condition, review text and date, and numerical features including review rating and useful counts.  In total, there are 215,063 examples, which is split into a training set of 75\% and a test set of 25\%. \\
    Settings & (1.0, 0.9, 0.1) \\
    Base Models & \tabitem \textbf{Ridge} - Ridge Regression (solver=`sage', random\_state=2020)  \\
                & \tabitem \textbf{MLP} - Multilayer Perceptron (Adam,lr=1e-3,manual\_seed=2020) \\
                & \tabitem \textbf{CNN} - Convolutional Neural Network (Adam,lr=1e-3,manual\_seed=2020) \\
    Benchmark & A public solution on Kaggle~\footnote{https://www.kaggle.com/stasian/predicting-review-scores-using-neural-networks} using two-layer Bidirectional-LSTM with Bahdanau Attention pooling before the prediction. \\
    Pre-processing & Ridge and MLP use average-pooling word embeddings from BERT-Base model of 768 dimensions, while CNN word embeddings are initialised from pre-trained word2vec of 300 dimensions. \\
    Run time & 1200s / epoch \\
    Observations & \\
    & \textbf{Learning Dynamics} as shown in \autoref{fig:learning_det}(d) and \autoref{fig:learning_sto}(d) \\
    & \tabitem The overall trend is improving for all models, which is characterised by the average performance (coloured in red). \\
    & \tabitem Ridge has performed consistently on the DR dataset. Despite that the performance is still the lowest as Ridge is the most simple one among the three base models. \\
    & \tabitem MLP's performance is the most unstable. We argue that it is due to the complexity of the task and MLP is easily  trapped in local minimums of the optimisation space. \\
    & \tabitem With the highest modelling capacity, CNN shows the best performance among the base models.  CNN is more stable than MLP though not as stable as Ridge. \\
    & \tabitem The sampling ratio is stabilised after several episodes and converged nicely over time. \\
    
    & \textbf{Comparison between RDS$^{DET}$ and RDS$^{STO}$} \\
	& \tabitem The optimisation is stable on both deterministic and stochastic reward mechanisms. \\
	& \tabitem Sampling ratio has become stable within the first 30 episodes.  Both converge to the expected ratio (0.75) over time.\\
	& \tabitem IID regularisation is preserved better with deterministic reward mechanism. \\
	
	& \textbf{Comparison between RDS and other methods as shown in \autoref{tab:dr_results}} \\
	& \tabitem The preset selection shows the highest performance on MLP; which may hint that the dataset was prepared with hold-out based on MLP model. However, it has the worst performance on Ridge and CNN. \\
	& \tabitem Random sampling technique shows sub-optimal results in all comparisons.  Notably, it yields the worst performance with the public solution. \\
	& \tabitem RDS has a good balance amongst all three base models. The performance of RDS variants on MLP is slightly lower than the performance of the preset selection. However, it yields higher results in Ridge, CNN, and Ensemble. \\
	& \tabitem RDS$^{STO}$ achieves the best performance for Ridge and CNN, while RDS$^{DET}$ achieving the highest results in the ensemble as well as on the public solution.  It suggests the effectiveness of using RDS techniques for data sampling to capture full model potentials. \\
	
	& \textbf{Public Benchmarking} \\
	& \tabitem The use of public solution has shown higher performance compared to the bare-bone ensemble of three base models.  It is consistent across all sampling strategies. \\
	& \tabitem The best performance is achieved by RDS$^{DET}$, which agrees with the ensemble as well. \\
    & \textbf{Summary} \\
	& The experimental results on Drug Review dataset suggest the effectiveness of reinforced sampling for model diversification.  The observations are also agreeable with other experiments.  \\
    \bottomrule
    \end{tabularx}%
  \label{tab:dr_notes}%
  \caption{Drug Review - Experiment Notes}
\end{table}

\begin{table}[H]
  \centering
    \begin{tabularx}{\textwidth}{lX}
    \toprule
    Experiment     & Notes \\
    \midrule
    Background & \textbf{MNIST}~\cite{lecun1998mnist} consists of 70,000 hand-written digits and is one of the most well-known datasets in the deep learning community. MNIST is selected for experiments since it represents very well for multi-class classification task on images. MNIST considered a balanced image classification data, and it is divided into 60K samples for training and 10K samples for testing. 
In this experiment, we use three base learners, including Logistic Regression (LR), Random Forest (RF), and Convolutional Neural Network (CNN). \\
    Settings & (1.0, 0.9, 0.1) \\
    % Base Models & \\
    
    Base Models & \tabitem \textbf{LR} - Logistic Regression (solver=`lbfgs')  \\
                & \tabitem \textbf{RF} - Random Forest (n\_estimators=50) \\
                & \tabitem \textbf{CNN} - Convolutional Neural Network (Adam, lr=0.01) \\
    
    Benchmark &  A high score solution on Kaggle for MNIST classification task~\cite{mnist:public:kaggle}. \\
    Pre-processing & We extract Histogram of Oriented Gradients (HOG) features for policy learner, LR, and RF algorithms. To reduce the dimension, we apply PCA with the n\_components of $0.95$. It means the number of components is selected such that the amount of variance that needs to be explained is greater than the specified percentage (i.e., 95\%). For CNN model, it runs directly on RGB codes with normalisation.\\
    Run time & On average 88s / epoch \\
    Observations & \\
    & \textbf{Learning Dynamics} as shown in \autoref{fig:learning_det}(c) and \autoref{fig:learning_sto}(c) \\
    & \tabitem LR, RF, CNN have performed consistently on the MNIST dataset. The upward trend can be observed during the agent learning across three base models. \\
    & \tabitem There is the balance in class ratios of the dataset. This observation is based on seeing the IID regulation approaches almost zero. \\
    & \textbf{Comparison between RDS$^{DET}$ and RDS$^{STO}$} \\
    & \tabitem The optimisation process is stable and gets better on both deterministic and stochastic reward mechanisms. \\
    & \tabitem CNN achieves the highest performance on both RDS$^{STO}$ setting and RDS$^{DET}$ settings. CNN can increasingly perform better suggests that the selected samples are well-represented and the model does not face class-imbalance issue. \\
    & \tabitem The stochastic reward mechanism has the same learning dynamics and achieves better performance in shorter time.\\
    & \textbf{Comparison between RDS and other methods as shown in \autoref{tab:mnist_results}} \\
    & \tabitem The preset selection gets the highest performance on RF. This might hint that the preset selection process was based on RF.\\
    & \tabitem The stratified setting gets the highest performance on LR. It showcases that stratified splitting approach is a strong baseline for balanced data.\\
    & \tabitem Both random and RDS$^{DET}$ show sub-optimal performance in all evaluation metrics. This hints that they might have the issue of imbalance selection between samples for training and testing. For RDS$^{DET}$, the majority voting was used for reward mechanism. It might be a reason to cause the problem of diversity. \\
	& \textbf{Public Benchmarking} \\
	& \tabitem The use of public solution has the best performance in comparison to the ensemble of three base models. The same evaluation metric is used for all sampling techniques. \\
	& \tabitem The RDS$^{STO}$ gets the best performance in public benchmarking. \\
    & \textbf{Summary} \\
    & \tabitem The MNIST experiment shows that the proposed approach can perform effectively on image classification task for model diversification. Given the fact that the preset setting of MNIST dataset was well-prepared for having good splitting in terms of both samples' similarity and classes' balance. However, our proposed RDS$^{STO}$ approach can effectively select a better split by showing that the public solution gets better performance in comparison to other splitting methods.\\
    \bottomrule
    \end{tabularx}%
  \label{tab:mnist_notes}%
  \caption{MNIST - Experiment Notes}
\end{table}

\begin{table}[H]
  \centering
    \begin{tabularx}{\textwidth}{lX}
    \toprule
    Experiment     & Notes \\
    \midrule
    Background & KLP was provided in the Kalapa credit scoring challenge \cite{klp2020credit}. It contains 50,000 profiles associated with good or bad labels. Each profile has 62 demographic and financial features. Originally, KLP is separated into 30,000 training and 20,000 testing examples with the strongly imbalanced data problem, i.e., only approximately 1.6\% the total number of profiles are labelled as 'good'. As another serious issue, over 40 fields have more than 30\% missing values.\\
    Settings & (1.0, 1.0, 40) \\
    Base Models & \tabitem \textbf{LR} - Logistic Regression (solver=`liblinear', random\_state=123)  \\
                & \tabitem \textbf{RF} - Random Forest (n\_estimators=64, random\_state=123) \\
                & \tabitem \textbf{MLP} - Multilayer Perceptron (Adam, lr=1e-3, manual\_seed=123) \\
    Benchmark & The 1st rank solution using Random Forest with WOE binnings and the number estimators of 767 \cite{klp2020sol} \\
    Pre-processing & Three text fields namely 'Province', 'Job', 'District' are extracted average-pooling word embeddings from a fine-tuned Fasttext model of 32 dimensions. Other fields are applied with traditional feature engineering techniques (e.g., MinMaxScaler, OrdinaryEncoder, dummy variables). \\
    Run time & On average 18s / epoch \\ 
    Observations & \\
    & \textbf{Learning Dynamics} as shown in \autoref{fig:learning_det}(b) and \autoref{fig:learning_sto}(b) \\
    & \tabitem LR is quite stable with higher performance compared against RF and MLP. \\
    & \tabitem RF is not always the most effective model for the classification task with the imbalanced and missing data problems. \\
    & \tabitem MLP shows a better performance without the constraint by the deterministic soft-voting reward mechanism.\\
    & \tabitem Regularisations work properly to force the sampling ratio converged to the expected value, and reduce the label distribution difference between the training and testing sets. \\
    & \textbf{Comparison between RDS$^{DET}$ and RDS$^{STO}$} \\
    & \tabitem There is a consistent upward trend on classification performance of the ensemble model for both reward mechanisms. \\
    & \tabitem The models are more relaxing on the learning process with the stochastic reward mechanism, hence they have better performance. \\
    & \tabitem The sampling ratio and IID regularisation are fluctuating within the first 30 episodes. Subsequently, they are more stable and converged after 450 episodes. A similar trend is observed for both reward mechanisms.   \\
    & \textbf{Comparison between RDS and other methods as shown in \autoref{tab:klp_results}} \\
    & \tabitem The best performance with the preset selection method is observed on LR. It implies that LR is the base model to split data in the credit scoring challenge. \\
    & \tabitem It shows sub-optimal performance for both random and stratified sampling methods on all evaluation metrics.  \\
    & \tabitem RDS variants outperform traditional sampling methods for almost individual or ensemble models. The performance of RDS on RF is slightly lower than the performance of the stratified selection due to the solid dependence of RS on feature distributions. \\
    & \textbf{Public Benchmarking} \\
    & \tabitem The public solution results in better performance compared against to the bare-bone ensemble of three base models across all sampling methods, except RDS$^{DET}$. \\
	& \tabitem The best performance is achieved by RDS$^{STO}$ \\
    & \textbf{Summary} \\
    & Experiments on KLP affirms the advantages of the reinforced sampling method for model diversification. With appropriate regularisation settings, the proposed method can help to effectively control sampling constraints, even if there are serious imbalanced and missing data problems. \\
    \bottomrule
    \end{tabularx}%
  \label{tab:klp_notes}%
  \caption{KLP - Experiment Notes}
\end{table}

\end{document}